\documentclass[11pt,a4paper]{article}
\usepackage[hyperref]{acl2020}
\usepackage{times}
\usepackage{latexsym}

\usepackage{microtype}

\usepackage{graphicx,color}
\usepackage{caption,todonotes}
\usepackage{amssymb}
\usepackage{amsmath}
\usepackage{multirow}
\usepackage{booktabs}
\usepackage{pifont}
\newcommand{\cmark}{\ding{51}}%
\newcommand{\xmark}{\ding{55}}%

\aclfinalcopy 

\title{From English to Code-Switching: Transfer Learning\\with Strong Morphological Clues}

\author{Gustavo Aguilar \and Thamar Solorio \\
  Department of Computer Science \\
  University of Houston\\
  Houston, TX 77204-3010 \\
  \texttt{\{gaguilaralas,  tsolorio\}@uh.edu}}

\date{}

\begin{document}
\maketitle
\begin{abstract}
Linguistic Code-switching (CS) is still an understudied phenomenon in natural language processing. The NLP community has mostly focused on monolingual and multi-lingual scenarios, but little attention has been given to CS in particular.
This is partly because of the lack of resources and annotated data, despite its increasing occurrence in social media platforms.
In this paper, we aim at adapting monolingual models to code-switched text in various tasks. 
Specifically, we transfer English knowledge from a pre-trained ELMo model to different code-switched language pairs (i.e., Nepali-English, Spanish-English, and Hindi-English) using the task of language identification. 
Our method, CS-ELMo, is an extension of ELMo with a simple yet effective position-aware attention mechanism inside its character convolutions. 
We show the effectiveness of this transfer learning step by outperforming multilingual BERT and homologous CS-unaware ELMo models and establishing a new state of the art in CS tasks, such as NER and POS tagging. 
Our technique can be expanded to more English-paired code-switched languages, providing more resources to the CS community.
\end{abstract}

\section{Introduction}

\begin{figure}[t!]
\renewcommand{\arraystretch}{1.2}
\centering
\resizebox{\linewidth}{!}{
\begin{tabular}{cc}
    \begin{tabular}{|l|}
    \hline
    \textbf{Hindi-English Tweet} \\
    \hline
        \textbf{Original:} \textit{Keep calm and keep} 
        \underline{kaam se kaam} !!!\textsubscript{\texttt{other}} \textit{\#office} \\
        \textit{\#tgif}
        \textit{\#nametag} \#buddha\textsubscript{\texttt{ne}} \textit{\#SouvenirFromManali \#keepcalm} \\
        \textbf{English:} Keep calm and mind your own business !!! \\
    \hline
    \end{tabular}
    \\\\
    \begin{tabular}{|l|}
    \hline
    \textbf{Nepali-English Tweet}  \\
    \hline
        \textbf{Original:} Youtube\textsubscript{\texttt{ne}} \underline{ma live re} ,\textsubscript{\texttt{other}} \underline{chalcha ki vanni aash} \\ 
        \underline{garam} !\textsubscript{\texttt{other}} \textit{Optimistic} .\textsubscript{\texttt{other}} \\
        \textbf{English:} They said Youtube live, let's hope it works! Optimistic. \\
    \hline
    \end{tabular} 
    \\\\
    \begin{tabular}{|l|}
    \hline
    \textbf{Spanish-English Tweet} \\
    \hline
        \textbf{Original:} @MROlvera06\textsubscript{\texttt{other}} @T11gRe\textsubscript{\texttt{other}} \textit{go too} \\
        cavenders\textsubscript{\texttt{ne}}
        \underline{y tambien ve a} @ElToroBoots\textsubscript{\texttt{ne}} \includegraphics[height=1em]{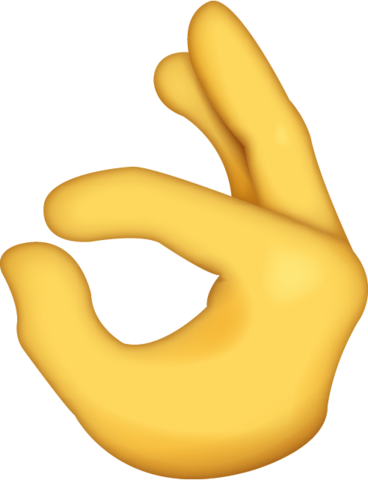}\textsubscript{\texttt{other}}  \\
        \textbf{English:} @MROlvera06 @T11gRe go to cavenders and \\
        also go to @ElToroBoots \includegraphics[height=1em]{images/emoji2.png}\\
    \hline
    \end{tabular} 
\end{tabular}
}
\caption{\label{fig:cs_examples}
Examples of code-switched tweets and their translations from the CS LID corpora for Hindi-English, Nepali-English and Spanish-English. 
The LID labels \texttt{ne} and \texttt{other} in subscripts refer to named entities and punctuation, emojis or usernames, respectively (they are part of the LID tagset).
English text appears in \textit{italics} and other languages are \underline{underlined}. }
\end{figure}

Although linguistic code-switching (CS) is a common phenomenon among multilingual speakers, it is still considered an understudied area in natural language processing. 
The lack of annotated data combined with the high diversity of languages in which this phenomenon can occur makes it difficult to strive for progress in CS-related tasks.
Even though CS is largely captured in social media platforms, it is still expensive to annotate a sufficient amount of data for many tasks and languages.
Additionally, not all the languages have the same incidence and predominance, making annotations impractical and expensive for every combination of languages.
Nevertheless, code-switching often occurs in language pairs that include English (see examples in Figure \ref{fig:cs_examples}).
These aspects lead us to explore approaches where English pre-trained models can be leveraged and tailored to perform well on code-switching settings.

In this paper, we study the CS phenomenon using English as a starting language to adapt our models to multiple code-switched languages, such as Nepali-English, Hindi-English and Spanish-English. 
In the first part, we focus on the task of language identification (LID) at the token level using ELMo \cite{peters-EtAl:2018:N18-1} as our reference for English knowledge. Our hypothesis is that English pre-trained models should be able to recognize whether a word belongs to English or not when such models are fine-tuned with code-switched text. 
To accomplish that, we introduce CS-ELMo, an extended version of ELMo that contains a position-aware hierarchical attention mechanism over ELMo's character n-gram representations.
These enhanced representations allow the model to see the location where particular n-grams occur within a word (e.g., affixes or lemmas) and to associate such behaviors with one language or another.\footnote{Note that there are more than two labels in the LID tagset, as explained in Section \ref{sec:dataset}.}
With the help of this mechanism, our models consistently outperform the state of the art on LID for Nepali-English \cite{solorio-etal-2014-overview}, Spanish-English \cite{molina-etal-2016-overview}, and Hindi-English \cite{mave-etal-2018-language}. Moreover, we conduct experiments that emphasize the importance of the position-aware hierarchical attention and the different effects that it can have based on the similarities of the code-switched languages. 
In the second part, we demonstrate the effectiveness of our CS-ELMo models by further fine-tuning them on tasks such as NER and POS tagging. 
Specifically, we show that the resulting models significantly outperform multilingual BERT and their homologous ELMo models directly trained for NER and POS tagging. Our models establish a new state of the art for Hindi-English POS tagging \cite{singh-etal-2018-twitter} and Spanish-English NER \cite{aguilar-etal-2018-named}.

Our contributions can be summarized as follows: 
1) we use transfer learning from models trained on a high-resource language (i.e., English) and effectively adapt them to the code-switching setting for multiple language pairs on the task of language identification;
2) we show the effectiveness of transferring a model trained for LID to downstream code-switching NLP tasks, such as NER and POS tagging, by establishing a new state of the art;
3) we provide empirical evidence on the importance of the enhanced character n-gram mechanism, which aligns with the intuition of strong morphological clues in the core of ELMo (i.e., its convolutional layers); and
4) our CS-ELMo model is self-contained, which allows us to release it for other researchers to explore and replicate this technique on other code-switched languages.\footnote{\url{http://github.com/RiTUAL-UH/cs_elmo}}

\section{Related Work}

Transfer learning has become more practical in the last years, making possible to apply very large neural networks to tasks where annotated data is limited \cite{howard-ruder-2018-universal, peters-EtAl:2018:N18-1, devlin2018bert}. 
CS-related tasks are good candidates for such applications, since they are usually framed as low-resource problems.
However, 
previous research on sequence labeling for code-switching mainly focused on traditional ML techniques because they performed better than deep learning models trained from scratch on limited data \cite{yirmibesoglu-eryigit-2018-detecting, al-badrashiny-diab-2016-lili}. 
Nonetheless, some researchers have recently shown promising results by using pre-trained monolingual embeddings for tasks such as 
NER \cite{trivedi-etal-2018-iit,winata-etal-2018-bilingual} and 
POS tagging \cite{soto-hirschberg-2018-joint, ball-garrette-2018-part}. 
Other efforts include the use of multilingual sub-word embeddings like \textit{fastText} \cite{bojanowski2017enriching} for LID \cite{mave-etal-2018-language}, 
and cross-lingual sentence embeddings for text classification like LASER \cite{schwenk-2018-filtering,SCHWENK18.658,schwenk-douze-2017-learning}, which is capable of handling code-switched sentences.
These results show the potential of pre-trained knowledge and they motivate our efforts to further explore transfer learning in code-switching settings.

Our work is based on ELMo \cite{peters-EtAl:2018:N18-1}, a large pre-trained language model that has not been applied to CS tasks before. 
We also use attention \cite{DBLP:journals/corr/BahdanauCB14} within ELMo's convolutions to adapt it to code-switched text. 
Even though attention is an effective and successful mechanism in other NLP tasks, 
the code-switching literature barely covers such technique \cite{sitaram2019survey}. 
\citet{wang-etal-2018-code} use a different attention method for NER, which is based on a gated cell that learns to choose appropriate monolingual embeddings according to the input text. Recently, \citet{winata-etal-2019-learning} proposed multilingual meta embeddings (MME) combined with self-attention \cite{transformers}. 
Their method establishes a state of the art on Spanish-English NER by heavily relying on monolingual embeddings for every language in the code-switched text.
Our model outperforms theirs by only fine-tuning a generic CS-aware model, without relying on task-specific designs.
Another contribution of our work are position embeddings, which have not been considered for code-switching either. 
These embeddings, combined with CNNs, have proved useful in computer vision \cite{DBLP:journals/corr/GehringAGYD17}; they help to localize non-spatial features extracted by convolutional networks within an image. 
We apply the same principle to code-switching: we argue that character n-grams without position information may not be enough for a model to learn the actual morphological aspects of the languages (e.g., affixes or lemmas). 
We empirically validate those aspects and discuss the incidence of such mechanism in our experiments.

\section{Methodology}
\label{sec:enhanced-char-ngrams}

\begin{figure*}
\centering
\includegraphics[width=\linewidth]{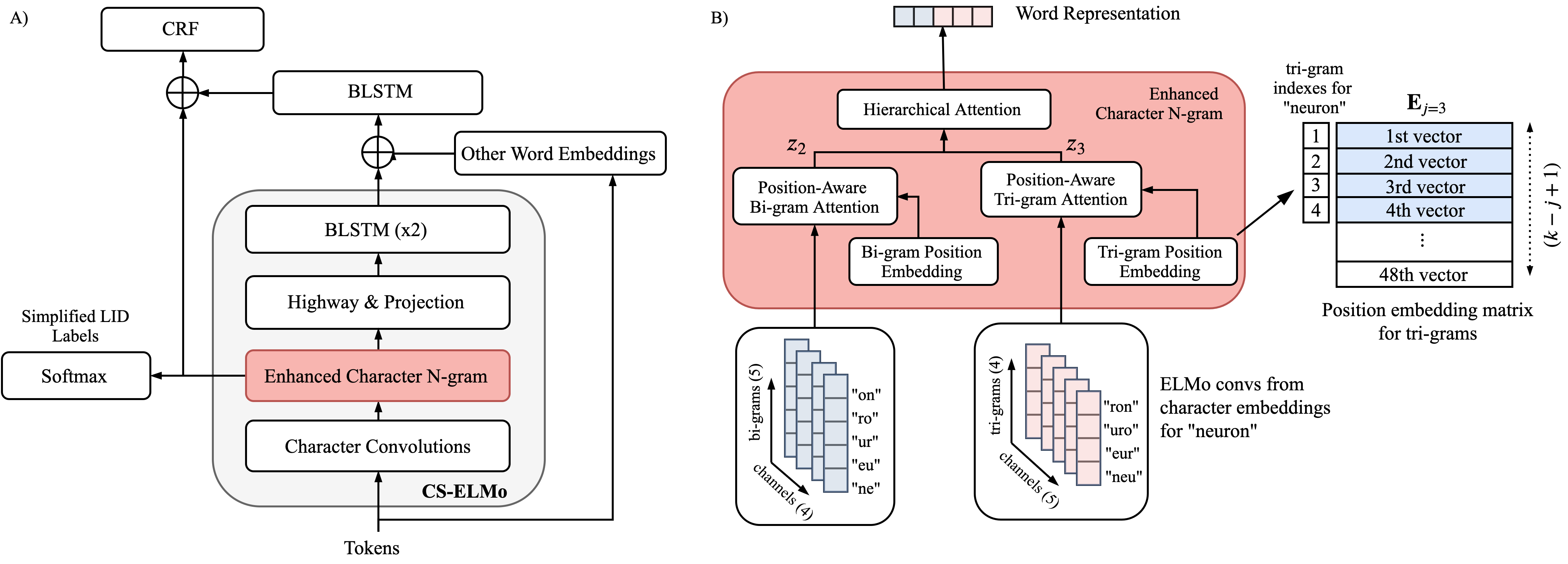}
\caption{ 
A) The left figure shows the overall model architecture, which contains CS-ELMo followed by BLSTM and CRF, and a secondary task with a softmax layer using a \textit{simplified} LID label set. The largest box describes the components of CS-ELMo, including the enhanced character n-gram module proposed in this paper. B) The right figure describes in detail the enhanced character n-gram mechanism inside CS-ELMo. The figure shows the convolutions of a word as input and a single vector representation as output.}
\label{fig:overall-model}
\end{figure*}

ELMo is a character-based language model that provides deep contextualized word representations \cite{peters-EtAl:2018:N18-1}.
We choose ELMo for this study for the following reasons:
1) it has been trained on a large amount of English data as a general-purpose language model and this aligns with the idea of having English knowledge as starting point; 
2) it extracts morphological information out of character sequences, which is essential for our case since certain character n-grams can reveal whether a word belongs to one language or another; and 
3) it generates powerful word representations that account for multiple meanings depending on the context.
Nevertheless, some aspects of the standard ELMo architecture could be improved to take into account more linguistic properties. In Section \ref{sec:position-aware}, we discuss these aspects and propose the position-aware hierarchical attention mechanism inside ELMo. In Section \ref{sec:sequential-tagger} and Section \ref{sec:multitask}, we describe our overall sequence labeling model and the training details, respectively.

\subsection{Position-Aware Hierarchical Attention}
\label{sec:position-aware}

ELMo convolves character embeddings in its first layers and uses the resulting convolutions to represent words. During this process, the convolutional layers are applied in parallel using different kernel sizes, which can be seen as character n-gram feature extractors of different orders.
The feature maps per n-gram order are max-pooled to reduce the dimensionality, and the resulting single vectors per n-gram order are concatenated to form a word representation. 
While this process has proven effective in practice, we notice the following shortcomings:
\begin{enumerate}
    \item Convolutional networks do not account for the positions of the character n-grams (i.e., convolutions do not preserve the sequential order), losing linguistic properties such as affixes.
    \item ELMo down-samples the outputs of its convolutional layers by max-pooling over the feature maps. However, this operation is not ideal to adapt to new morphological patterns from other languages as the model tends to discard patterns from languages other than English.
\end{enumerate}
To address these aspects, we introduce CS-ELMo, an extension of ELMo that incorporates a position-aware hierarchical attention mechanism that enhances ELMo's character n-gram representations. 
This mechanism is composed of three elements: position embeddings, position-aware attention, and hierarchical attention. Figure \ref{fig:overall-model}A describes the overall model architecture, and Figure \ref{fig:overall-model}B details the components of the enhanced character n-gram mechanism.

\paragraph{Position embeddings.} 
Consider the word $\mathbf{x}$ of character length $l$, whose character n-gram vectors are  $(x_1, x_2, \dots, x_{l-j+1})$ for an n-gram order $j \in \{1, 2, \dots, n\}$.\footnote{ELMo has seven character convolutional layers, each layer with a kernel size from one to seven characters  ($n=7$).} The n-gram vector $x_i \in \mathbb{R}^{c}$ is the output of a character convolutional layer, where $c$ is the number of output channels for that layer. 
Also, consider $n$ position embedding matrices, one per n-gram order, $\{\mathbf{E}_1, \mathbf{E}_2, \dots, \mathbf{E}_n\}$ defined as $\mathbf{E}_j \in \mathbb{R}^{(k - j + 1) \times e}$ where $k$ is the maximum length of characters in a word (note that $l \le k$), $e$ is the dimension of the embeddings and $j$ is the specific n-gram order. 
Then, the position vectors for the sequence $\mathbf{x}$ are defined by $\mathbf{p} = (p_1, p_2, \dots, p_{l-j+1})$ where $p_i \in \mathbb{R}^{e}$ is the $i$-th vector from the position embedding matrix $\mathbf{E}_j$. We use $e = c$ to facilitate the addition of the position embeddings and the n-gram vectors.\footnote{ELMo varies the output channels per convolutional layer, so the dimensionality of $\mathbf{E}_j$ varies as well.} 
Figure \ref{fig:overall-model}B illustrates the position embeddings for bi-grams and tri-grams. 

\paragraph{Position-aware attention.} 
Instead of down-sampling with the max-pooling operation, we use an attention mechanism similar to the one introduced by \citet{DBLP:journals/corr/BahdanauCB14}. The idea is to concentrate mass probability over the feature maps that capture the most relevant n-gram information along the word, while also considering positional information. At every individual n-gram order, our attention mechanism uses the following equations:
\begin{align} 
	u_i =&~ v^\intercal ~\mathrm{tanh}(\mathrm{W}_x x_i + p_i + b_x) \label{eq:att-raw-scores} \\
	\alpha_i =&~ \frac{\mathrm{exp}(u_i)}{\sum^{N}_{j=1}\mathrm{exp}(u_j)}, ~~~\text{s.t.}~\sum_{i=1} \alpha_i = 1 \label{eq:att-alpha} \\
	z =& \sum_{i=1}{\alpha_i x_i} \label{eq:att-result}
\end{align}
\noindent where $\mathrm{W}_x \in \mathbb{R}^{a \times c}$ is a projection matrix, $a$ is the dimension of the attention space, $c$ is the number of channels for the n-gram order $j$, and $p_i$ is the position embedding associated to the $x_i$ n-gram vector. 
$v \in \mathbb{R}^{a}$ is the vector that projects from the attention space to the unnormalized scores, and $\alpha_i$ is a scalar that describes the attention probability associated to the $x_i$ n-gram vector. 
$z$ is the weighted sum of the input character n-gram vectors and the attention probabilities, which is our down-sampled word representation for the n-gram order $j$.
Note that this mechanism is used independently for every order of n-grams resulting in a set of $n$ vectors $\{z_1, z_2, \dots, z_n\}$ from Equation \ref{eq:att-result}. This allows the model to capture relevant information across individual n-grams before they are combined (i.e., processing independently all bi-grams, all tri-grams, etc.). 

\paragraph{Hierarchical attention.} 
With the previous mechanisms we handle the problems aforementioned. That is, we have considered positional information as well as the attention mechanism to down-sample the dimensionality. These components retrieve one vector representation per n-gram order per word. While ELMo simply concatenates the n-gram vectors of a word, we decide to experiment with another layer of attention that can prioritize n-gram vectors across all the orders. 
We use a similar formulation to Equations \ref{eq:att-raw-scores} and \ref{eq:att-result}, except that we do not have $p_i$, and instead of doing the weighted sum, we concatenate the weighted inputs. 
This concatenation keeps the original dimensionality expected in the upper layers of ELMo, while it also emphasizes which n-gram order should receive more attention.

\subsection{Sequence Tagging}
\label{sec:sequential-tagger}

We follow \citet{peters-EtAl:2018:N18-1} to use ELMo for sequence labeling. 
They reported state-of-the-art performance on NER by using ELMo followed by a bidirectional LSTM layer and a linear-chain conditional random field (CRF). We use this architecture as a backbone for our model (see Figure \ref{fig:overall-model}A), but we add some modifications. 
The first modification is the concatenation of static English word embeddings to ELMo's word representation, such as Twitter \cite{pennington-etal-2014-glove} and \textit{fastText} \cite{bojanowski2017enriching} embeddings similar to \citet{howard-ruder-2018-universal} and \citet{mave-etal-2018-language}. The idea is to enrich the context of the words by providing domain-specific embeddings and sub-word level embeddings. 
The second modification is the concatenation of the enhanced character n-gram representation with the input to the CRF layer. This emphasizes even further the extracted morphological patterns, so that they are present during inference time for the task at hand (i.e., not only LID, but also NER and POS tagging). 
The last modification is the addition of a secondary task on a \textit{simplified}\footnote{The LID label set uses eight labels (\texttt{lang1}, \texttt{lang2}, \texttt{ne}, \texttt{mixed}, \texttt{ambiguous}, \texttt{fw}, \texttt{other}, and \texttt{unk}), but for the \textit{simplified} LID label set, we only consider three labels (\texttt{lang1}, \texttt{lang2} and \texttt{other}) to predict only based on characters.} language identification label scheme (see Section \ref{sec:dataset} for more details), which only uses the output of the enhanced character n-gram mechanism. Intuitively, this explicitly forces the model to associate morphological patterns (e.g., affixes, lemmas, etc.) to one or the other language.

\subsection{Multi-Task Training}
\label{sec:multitask}

We train the model by minimizing the negative log-likelihood loss of the CRF classifier. 
Additionally, we force the model to minimize a secondary loss over the \textit{simplified} LID label set by only using the morphological features from the enhanced character n-gram mechanism (see the softmax layer in Figure \ref{fig:overall-model}A). The overall loss $\mathcal{L}$ of our model is defined as follows:
\begin{align}
	\mathcal{L}_{task_t} =~ -\frac{1}{N}\sum_i^N y_{i} ~log~p(y_{i}|\Theta) \label{eq:general_nllloss}\\
	\mathcal{L} =~ \mathcal{L}_{task_1} + \beta\mathcal{L}_{task_2} + \lambda \sum^{|\Theta|}_k{w_k^2} 
\end{align}
\noindent where $\mathcal{L}_{task_1}$ and $\mathcal{L}_{task_2}$ are the negative log-likelihood losses conditioned by the model parameters $\Theta$ as defined in Equation \ref{eq:general_nllloss}. $\mathcal{L}_{task_1}$ is the loss of the primary task (i.e., LID, NER, or POS tagging), whereas $\mathcal{L}_{task_2}$ is the loss for the \textit{simplified} LID task weighted by $\beta$ to smooth its impact on the model performance. 
Both losses are the average over $N$ tokens.\footnote{While Equation \ref{eq:general_nllloss} is formulated for a given sentence, in practice $N$ is the number of tokens in a batch of sentences.}
The third term provides $\ell_2$ regularization, and $\lambda$ is the penalty weight.\footnote{We exclude the CRF parameters in this term.} 

\section{Datasets}
\label{sec:dataset}

\paragraph{Language identification.} 
We experiment with code-switched data for Nepali-English, Spanish-English, and Hindi-English. The first two datasets were collected from Twitter, and they were introduced at the 
Computational Approaches to Linguistic Code-Switching
(CALCS) workshops in 2014 and 2016 \cite{solorio-etal-2014-overview,molina-etal-2016-overview}. The Hindi-English dataset contains Twitter and Facebook posts, and it was introduced by \citet{mave-etal-2018-language}.
These datasets follow the CALCS label scheme, which has eight labels: 
\texttt{lang1} (English), \texttt{lang2} (Nepali, Spanish, or Hindi), \texttt{mixed}, \texttt{ambiguous}, \texttt{fw}, \texttt{ne}, \texttt{other}, and \texttt{unk}. We show the distribution of \texttt{lang1} and \texttt{lang2} in Table \ref{tab:dataset_splits}. 
Moreover, we add a second set of labels using a \textit{simplified} LID version of the original CALCS label set. The simplified label set uses \texttt{lang1}, \texttt{lang2}, and \texttt{other}. We use this 3-way token-level labels in the secondary loss of our model where only morphology, without any context, is being exploited. 
This is because we are interested in predicting whether a word's morphology is associated to English more than to another language (or vice versa), 
instead of whether, for example, its morphology describes a named entity (\texttt{ne}).


\begin{table}[t!]
\centering
\resizebox{0.95\linewidth}{!}{
\begin{tabular}{llrrrr} 
\toprule
\textbf{Corpus} & \textbf{Split} & \textbf{Posts} & \textbf{Tokens} & \textbf{Lang1} & \textbf{Lang2} \\
\midrule
\multirow{3}{*}{Nep-Eng}
    & Train & 8,494     & 123,959   & 38,310    & 51,689 \\
    & Dev   & 1,499     &  22,097   &  7,173    &  9,008 \\
    & Test  & 2,874     &  40,268   & 12,286    & 17,216 \\
\midrule
\multirow{3}{*}{Spa-Eng}
    & Train & 11,400 & 139,539 & 78,814	& 33,709 \\
    & Dev   & 3,014  & 33,276  & 16,821 & 8,652     \\
    & Test  & 10,716 & 121,446 & 16,944 & 77,047  \\
\midrule
\multirow{3}{*}{Hin-Eng}
    & Train & 5,045	& 100,337 & 57,695 & 20,696 \\
    & Dev   &   891 &  16,531 &  9,468 &  3,420 \\
    & Test  & 1,485 &  29,854 &	17,589 &  5,842 \\ 	   
\bottomrule
\end{tabular}
}
\caption{The distribution of the LID datasets according to the CALCS LID label set. The label \texttt{lang1} refers to English and \texttt{lang2} is either Nepali, Spanish or Hindi depending on the corpus. The full label distribution is in Appendix \ref{app:lid-full-distribution}.}
\label{tab:dataset_splits}
\end{table}

\paragraph{Part-of-speech tagging.}
\citet{singh-etal-2018-twitter} provide 1,489 tweets (33,010 tokens) annotated with POS tags.
The labels are annotated using the universal POS tagset proposed by \citet{petrov-etal-2012-universal} with the addition of two labels: \texttt{PART\_NEG} and \texttt{PRON\_WH}. This dataset does not provide training, development, or test splits due to the small number of samples. Therefore, we run 5-fold cross validations and report the average scores.

\begin{table*}[t!]
\centering
\small
\setlength{\tabcolsep}{7pt}
\renewcommand{\arraystretch}{0.93}
\begin{tabular}{llllllll}
\toprule
\multirow{2}{*}{\textbf{\textbf{Exp ID}}} & 
\multirow{2}{*}{\textbf{\textbf{Experiment}}} &
\multicolumn{2}{l}{\textbf{Nepali-English}}	& 
\multicolumn{2}{l}{\textbf{Spanish-English}} & 
\multicolumn{2}{l}{\textbf{Hindi-English}}	 \\
~ & ~ & Dev & Test & Dev & Test & Dev & Test \\
\midrule
\multicolumn{2}{l}{\textit{\textbf{Approach 1} (Baseline models)}} \\
\midrule
Exp 1.1 & ELMo	                        & 96.192	& 95.700	& 95.508	& 96.363	& 95.997	& 96.420 \\
Exp 1.2 & ELMo + BLSTM + CRF            & \bf96.320	& 95.882	& 95.615	& \bf96.748	& \bf96.545	& \bf96.717 \\
Exp 1.3 & ML-BERT	                    & 95.436    & \bf96.571 & \bf96.212 & 96.212 & 95.924    & 96.440     \\
\midrule
\multicolumn{8}{l}{\textit{\textbf{Approach 2} (Upon Exp 1.2)}} \\
\midrule
Exp 2.1 & Attention on each n-gram			        & 96.413    &	96.771 &	95.952 &	96.519 &	96.579 &	96.069 \\
Exp 2.2 & Position-aware attention on each n-gram   & 96.540    &	96.640 &	95.994 &	\bf96.791 &	96.629 &	96.141 \\
Exp 2.3 & Position-aware hierarchical attention		& \bf96.582 &\bf96.798 &	\bf96.072 &	96.692 &	\bf96.705 &	\bf96.186 \\
\midrule
\multicolumn{2}{l}{\textit{\textbf{Approach 3} (Upon Exp 2.3) }} \\
\midrule
Exp 3.1 & Concatenating character n-grams at the top & 96.485    &	96.761  &	96.033 &	96.775 &	96.665 &	96.188 \\
Exp 3.2 & Adding simplified LID (secondary) task     & 96.612    &	96.734  &	96.051 &	96.932 &	96.565 &	96.215 \\
Exp 3.3 & Adding static word embeddings              & \underline{\bf96.879} &	\underline{\bf97.026} &	\underline{\bf96.757} &	\underline{\bf97.532} &	\underline{\bf96.776} &	\underline{\bf97.001} \\
\midrule
\multicolumn{2}{l}{\textit{\textbf{Comparison:} Previous best published results}} \\
\midrule
\multicolumn{2}{l}{\citet{mave-etal-2018-language}} & - & - & 96.510 & 97.060 & 96.6045 & 96.840 \\
\bottomrule
\end{tabular}
\caption{
The results of incremental experiments on each LID dataset. 
The scores are calculated using the weighted F-1 metric across the eight LID labels from CALCS. 
Within each column, the best score in each block is in \textbf{bold}, and the best score for the whole column is \underline{underlined}.
Note that development scores from subsequent experiments (e.g., Exp 2.2 and 2.3) are statistically significant with $p < 0.02$. 
}
\label{tab:lid-experiments}
\end{table*}

\paragraph{Named entity recognition.}
We use the Spanish-English NER corpus introduced in the 2018 CALCS competition \cite{aguilar-etal-2018-named}, which contains a total of 67,223 tweets with 808,663 tokens. The entity types are \texttt{person}, \texttt{organization}, \texttt{location}, \texttt{group}, \texttt{title}, \texttt{product}, \texttt{event}, \texttt{time}, and \texttt{other}, and the labels follow the BIO scheme. We used the fixed training, development, and testing splits provided with the datasets to benchmark our models.

Importantly, Hindi and Nepali texts in these datasets appear transliterated using the English alphabet (see Figure \ref{fig:cs_examples}). 
The lack of a standardized transliteration process leads code-switchers to employ mostly ad-hoc phonological rules that conveniently use the English alphabet when they write in social media. 
This behavior makes the automated processing of these datasets more challenging because it excludes potentially available resources in the original scripts of the languages.

\section{Experiments}

\begin{table}[t!]
\resizebox{\linewidth}{!}{
    \begin{tabular}{lllll} 
        \toprule
        \textbf{Corpus} & \textbf{LID System} & \textbf{Lang1} & \textbf{Lang2} & \textbf{WA F1} \\
        \midrule
        \multirow{3}{*}{Spa-Eng}
            & \citeauthor{al-badrashiny-diab-2016-lili}     & 88.6      & 96.9        & 95.2 \\
            & \citeauthor{jain-bhat-2014-language}          & 92.3      & 96.9        & 96.0 \\
            & \citeauthor{mave-etal-2018-language}          & 93.184    & 98.118      & 96.840 \\
            & Ours (Exp 3.3)                                & \bf94.411 & \bf98.532   & \underline{\bf97.789} \\
        \midrule
        \multirow{2}{*}{Hin-Eng}
            & \citeauthor{mave-etal-2018-language}          & 98.241    & 95.657    & 97.596 \\
            & Ours (Exp 3.3)                                & \bf98.372 & \bf95.750 & \underline{\bf97.718} \\
        \midrule
        \multirow{2}{*}{Nep-Eng}
            & \citeauthor{al-badrashiny-diab-2016-lili}     & 97.6      & \bf97.0      & 97.3 \\
            & Ours (Exp 3.3)                                & \bf98.124    & 95.170    & \underline{\bf97.387} \\
        \bottomrule
    \end{tabular}
}
\caption{Comparison of our best models with the best published scores for language identification. 
Scores are calculated with the F1 metric, and WA F1 is the weighted average F1 between both languages.}
\label{tab:lid-sota}
\end{table}

We describe our experiments for LID in Section \ref{sec:exp-lid}, including insights of the optimized models. 
In Section \ref{sec:exp-transfer-learning}, the optimized LID models are further fine-tuned on downstream NLP tasks, such as NER and POS tagging, to show the effectiveness of our preliminary CS adaptation step.
We test for statistical significance across our incremental experiments following \citet{dror18significance}, and we report p-values below $0.02$ for LID. We discuss hyperparameters and fine-tuning details in Appendix \ref{app:sec:fine-tuning}.

\subsection{Language Identification}
\label{sec:exp-lid}

\paragraph{Approach 1.} 
We establish three strong baselines using a vanilla ELMo (Exp 1.1), ELMo combined with BLSTM and CRF (Exp 1.2) as suggested by \citet{peters-EtAl:2018:N18-1}, and a multilingual BERT (Exp 1.3) provided by \citet{devlin2018bert}. We experiment with frozen weights for the core parameters of ELMo and BERT, but we find the best results when the full models are fine-tuned, which we report in Table \ref{tab:lid-experiments}. 

\paragraph{Approach 2.} 
In the second set of experiments, we add the components of our mechanism upon ELMo combined with BLSTM and CRF (Exp 1.2). We start by replacing the max-pooling operation with the attention layer at every individual n-gram order in Exp 2.1. In Exp 2.2, we incorporate the position information. The third experiment, Exp 2.3, adds the hierarchical attention across all n-gram order vectors. It is worth noting that we experiment by accumulating consecutive n-gram orders, and we find that the performance stops increasing when $n > 3$. Intuitively, this can be caused by the small size of the datasets since n-gram features of greater order are infrequent and would require more data to be trained properly. We apply our mechanism for n-gram orders in the set \{1, 2, 3\}, which we report in Table \ref{tab:lid-experiments}.

\paragraph{Approach 3.} 
For the third set of experiments, we focus on emphasizing the morphological clues extracted by our mechanism (Exp 2.3).
First, in Exp 3.1, we concatenate the enhanced character n-grams with their corresponding word representation before feeding the input to the CRF layer. 
In Exp 3.2, we add the secondary task over the previous experiment to force the model to predict the simplified LID labels by only using the morphological clues (i.e., no context is provided). Finally, in Exp 3.3, we add static word embeddings that help the model to handle social media style and domain-specific words. 

We achieve the best results on Exp 3.3, which outperforms both the baselines and the previous state of the art on the full LID label scheme (see Table \ref{tab:lid-experiments}).
However, to compare with other work, we also calculate the average of the weighted F1 scores over the labels \texttt{lang1} and \texttt{lang2}. Table \ref{tab:lid-sota} shows a comparison of our results and the previous state of the art. 
Note that, for Spanish-English and Hindi-English, the gap of improvement is reasonable, considering that similar gaps in the validation experiments are statistically significant. 
In contrast, in the case of Nepali-English, we cannot determine whether our improvement is marginal or substantial since the authors only provide one decimal in their scores. 
Nevertheless, \citet{al-badrashiny-diab-2016-lili} use a CRF with hand-crafted features \cite{al-badrashiny-diab-2016-lili}, while our approach does not require any feature engineering. 

\subsection{POS Tagging and NER}
\label{sec:exp-transfer-learning}

\begin{table}[t!]
\centering
\resizebox{\linewidth}{!}{
\begin{tabular}{lll} 
\toprule
\textbf{POS System} & \textbf{Dev F1} & \textbf{Test F1} \\ 
\midrule
ML-BERT                                      & 86.84 	& 84.70     \\
ELMo + BLSTM + CRF                           & 87.42	& 88.12     \\
Prev. SOTA \cite{singh-etal-2018-twitter} & -        & 90.20     \\
\midrule
\multicolumn{3}{l}{\textit{Architecture: CS-ELMo + BLSTM + CRF}} \\
Exp 4.1: No CS knowledge         & 87.02    & 87.96      \\ 
Exp 4.2: CS knowledge frozen     & 89.55    & 89.92      \\ 
Exp 4.3: CS knowledge trainable  & \bf90.37    & \bf91.03  \\ 
\bottomrule
\end{tabular}
}
\caption{
The F1 scores on POS tagging for the Hindi-English dataset. 
CS knowledge means that the CS-ELMo architecture (see Figure \ref{fig:overall-model}A) has been adapted to code-switching by using the LID task.
}
\label{tab:pos-sota}
\end{table}

\begin{table}[t!]
\centering
\resizebox{\linewidth}{!}{
\begin{tabular}{lll} 
\toprule
\textbf{NER System} & \textbf{Dev F1} & \textbf{Test F1} \\ 
\midrule
ML-BERT                                         & 61.11                 & 64.56    \\
ELMo + BLSTM + CRF                              & 59.91                 & 63.53    \\
Best at CALCS \cite{trivedi-etal-2018-iit}  & -                     & 63.76    \\
Prev. SOTA \cite{winata-etal-2019-learning}  & -                     & 66.63    \\
\midrule
\multicolumn{3}{l}{\textit{Architecture: CS-ELMo + BLSTM + CRF}} \\
Exp 5.1: No CS knowledge        & 62.59	        & 66.30 \\
Exp 5.2: CS knowledge frozen    & \textbf{64.39}	& \textbf{67.96} \\
Exp 5.3: CS knowledge trainable & 64.28	        & 66.84 \\
\bottomrule

\end{tabular}
}
\caption{
The F1 scores on the Spanish-English NER dataset. 
CS knowledge means that the CS-ELMo architecture (see Figure \ref{fig:overall-model}A) has been adapted to code-switching by using the LID task.
}
\label{tab:ner-sota}
\end{table}

We use LID to adapt the English pre-trained knowledge of ELMo to the code-switching setting, effectively generating CS-ELMo. 
Once this is achieved, we fine-tune the model on downstream NLP tasks such as POS tagging and NER. 
In this section, our goal is to validate whether the CS-ELMo model can improve over vanilla ELMo, multilingual BERT, and the previous state of the art for both tasks.
More specifically, we use our best architecture (Exp 3.3) from the LID experiments 
1) without the code-switching adaptation,
2) with the code-switching adaptation and only retraining the inference layer, and
3) with the code-switching adaptation and retraining the entire model. 

\paragraph{POS tagging experiments.} 
Table \ref{tab:pos-sota} shows our experiments on POS tagging using the Hindi-English dataset. 
When we compare our CS-ELMO + BLSTM + CRF model without CS adaptation (Exp 4.1) against the baseline (ELMo + BLSTM + CRF), the performance remains similar. 
This suggests that our enhanced n-gram mechanism can be added to ELMo without impacting the performance even if the model has not been adapted to CS. 
Slightly better performance is achieved when the CS-ELMo has been adapted to code-switching, and only the BLSTM and CRF layers are retrained (Exp 4.2). This result shows the convenience of our model since small improvements can be achieved faster by leveraging the already-learned CS knowledge while avoiding to retrain the entire model. 
Nevertheless, the best performance is achieved by the adapted CS-ELMO + BLSTM + CRF when retraining the entire model (Exp 4.3). Our results are better than the baselines and the previous state of the art.

\begin{figure*}[t!]
\centering
\includegraphics[width=\linewidth]{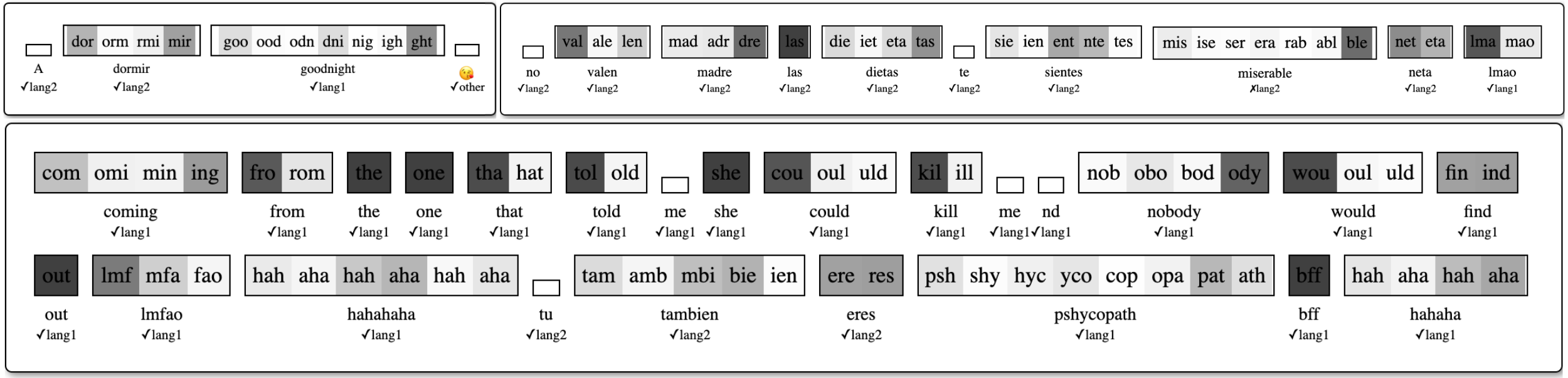}
\caption{ 
Visualization of the tri-gram attention weights for the 2016 Spanish-English LID dataset. 
The boxes contain the tri-grams of the word below them along with the right (\cmark) or wrong (\xmark) predictions by the model.
}
\label{fig:attention-visualization}
\end{figure*}

Interestingly, our model improves over multilingual BERT, which is a powerful and significantly bigger model in terms of parameters. 
Our intuition is that this is partly due to the word-piece tokenization process combined with the transliteration of Hindi. 
The fact that we use the multilingual version of BERT does not necessarily help to handle transliterated Hindi, since Hindi is only present in BERT's vocabulary with the Devanagari script. 
Indeed, we notice that in some tweets, the original number of tokens was almost doubled by the greedy tokenization process in BERT. 
This behavior tends to degrade the syntactic and semantic information captured in the original sequence of tokens. 
In contrast, ELMo generates contextualized word representations out of character sequences, which makes the model more suitable to adapt to the transliteration of Hindi.

\paragraph{NER experiments.} 
Table \ref{tab:ner-sota} contains our experiments on NER using the 2018 CALCS Spanish-English dataset. 
Exp 5.1 shows that the enhanced n-gram mechanism can bring improvements over the ELMo + BLSTM + CRF baseline, even though the CS-ELMo has not been adapted to the code-switching setting. 
However, better results are achieved when the CS-ELMo model incorporates the code-switching knowledge in both Exp 5.2 and 5.3. 
Unlike the POS experiments 4.2 and 4.3, fixing the parameters of CS-ELMo model yields better results than updating them during training.
Our intuition is that, in the NER task, the model needs the context of both languages to recognize entities within the sentences, and having the code-switching knowledge fixed becomes beneficial. 
Also, by freezing the CS-ELMo model, we can accelerate training because there is no backpropagation for the CS-ELMo parameters, which makes our code-switching adapatation very practical for downstream tasks.

\section{Analysis}
\label{sec:analysis}

\noindent\textbf{Position embeddings.} 
Localizing n-grams within a word is an important contribution of our method. 
We explore this mechanism by using our fine-tuned CS-ELMo to predict the \textit{simplified} LID labels on the validation set from the secondary task (i.e., the predictions solely rely on morphology) in two scenarios. 
The first one uses the position embeddings corresponding to the actual place of the character n-gram, whereas the second one chooses position embeddings randomly. 
We notice a consistent decay in performance across the language pairs, and a variation in the confidence of the predicted classes. 
The most affected language pair is Spanish-English, with an average difference of 0.18 based on the class probability gaps between both scenarios. 
In contrast, the probability gaps in Hindi-English and Nepali-English are substantially smaller; 
their average differences are 0.11 and 0.09, respectively.

\paragraph{Position distribution.} 
Considering the previous analysis and the variations in the results, 
we gather insights of the attention distribution according to their n-gram positions (see position-aware attention in Section \ref{sec:position-aware}). 
Although the distribution of the attention weights across n-gram orders mostly remain similar along the positions for all language pairs, Spanish-English has a distinctive concentration of attention at the beginning and end of the words. 
This behavior can be caused by the differences and similarities between the language pairs. For Spanish-English, the model may rely on inflections of similar words between the languages, such as affixes. 
On the other hand, transliterated Hindi and Nepali tend to have much less overlap with English words (i.e., words with few characters can overlap with English words), making the distinction more spread across affixes and lemmas.

\paragraph{Attention analysis.} 
Figure \ref{fig:attention-visualization} shows the tri-gram attention weights in the Spanish-English LID dataset. 
The model is able to pick up affixes that belong to one or the other language. 
For instance, the tri-gram \textit{-ing} is commonly found in English at the end of verbs in present progressive, like in the word \textit{com\textbf{ing}} from the figure, but it also appears in Spanish at different places (e.g., \textit{\textbf{ing}eniero}) making the position information relevant.
On the contrary, the tri-grams \textit{aha} and \textit{hah} from the figure do not seem to rely on position information because the attention distribution varies along the words. See more examples in Appendix \ref{sec:appendix_hi_en_viz}.

\paragraph{Error analysis.} 
Morphology is very useful for LID, but it is not enough when words have similar spellings between the languages. 
We inspect the predictions of the model, and find cases where, for example, \textit{miserable} is gold-labeled as \texttt{ambiguous} but the model predicts a language (see the top-right tweet in Figure \ref{fig:attention-visualization}).
Although we find similar cases for Nepali-English and Hindi-English, it mostly happens for words with few characters (e.g., \textit{me}, \textit{to}, \textit{use}).
The model often gets such cases mislabeled due to the common spellings in both languages. Although this should be handled by context, our contribution relies more on morphology than contextualization, which we leave for future work.

\section{Conclusion and Future Work}

We present a transfer learning method from English to code-switched languages using the LID task. 
Our method enables large pre-trained models, such as ELMo, to be adapted to code-switching settings while taking advantage of the pre-trained knowledge. 
We establish new state of the art on LID for Nepali-English, Spanish-English, and Hindi-English. 
Additionally, we show the effectiveness of our CS-ELMo model by further fine-tuning it for NER and POS tagging. 
We outperform multilingual BERT and homologous ELMo models on Spanish-English NER and Hindi-Enlgish POS tagging.
In our ongoing research, we are investigating the expansion of this technique to language pairs where English may not be involved.

\section*{Acknowledgements}

This work was supported by the National Science Foundation (NSF) on the grant \#1910192. 
We thank Deepthi Mave for providing general statistics of the code-switching datasets and Mona Diab for insightful discussions on the topic.

\bibliography{acl2020}

\begin{thebibliography}{29}
\expandafter\ifx\csname natexlab\endcsname\relax\def\natexlab#1{#1}\fi

\bibitem[{Aguilar et~al.(2018)Aguilar, AlGhamdi, Soto, Diab, Hirschberg, and
  Solorio}]{aguilar-etal-2018-named}
Gustavo Aguilar, Fahad AlGhamdi, Victor Soto, Mona Diab, Julia Hirschberg, and
  Thamar Solorio. 2018.
\newblock \href {https://www.aclweb.org/anthology/W18-3219} {{Named Entity
  Recognition on Code-Switched Data: Overview of the CALCS 2018 Shared Task}}.
\newblock In \emph{{Proceedings of the Third Workshop on Computational
  Approaches to Linguistic Code-Switching}}, pages 138--147, Melbourne,
  Australia. Association for Computational Linguistics.

\bibitem[{Al-Badrashiny and Diab(2016)}]{al-badrashiny-diab-2016-lili}
Mohamed Al-Badrashiny and Mona Diab. 2016.
\newblock \href {https://www.aclweb.org/anthology/C16-1115} {{{LILI}: A Simple
  Language Independent Approach for Language Identification}}.
\newblock In \emph{Proceedings of {COLING} 2016, the 26th International
  Conference on Computational Linguistics: Technical Papers}, pages 1211--1219,
  Osaka, Japan. The COLING 2016 Organizing Committee.

\bibitem[{Bahdanau et~al.(2015)Bahdanau, Cho, and
  Bengio}]{DBLP:journals/corr/BahdanauCB14}
Dzmitry Bahdanau, Kyunghyun Cho, and Yoshua Bengio. 2015.
\newblock \href {http://arxiv.org/abs/1409.0473} {{Neural Machine Translation
  by Jointly Learning to Align and Translate}}.
\newblock In \emph{3rd International Conference on Learning Representations,
  {ICLR} 2015, San Diego, CA, USA, May 7-9, 2015, Conference Track
  Proceedings}.

\bibitem[{Ball and Garrette(2018)}]{ball-garrette-2018-part}
Kelsey Ball and Dan Garrette. 2018.
\newblock \href {https://www.aclweb.org/anthology/D18-1347} {{Part-of-Speech
  Tagging for Code-Switched, Transliterated Texts without Explicit Language
  Identification}}.
\newblock In \emph{Proceedings of the 2018 Conference on Empirical Methods in
  Natural Language Processing}, pages 3084--3089, Brussels, Belgium.
  Association for Computational Linguistics.

\bibitem[{Bojanowski et~al.(2017)Bojanowski, Grave, Joulin, and
  Mikolov}]{bojanowski2017enriching}
Piotr Bojanowski, Edouard Grave, Armand Joulin, and Tomas Mikolov. 2017.
\newblock \href {https://doi.org/10.1162/tacl_a_00051} {{Enriching Word Vectors
  with Subword Information}}.
\newblock \emph{Transactions of the Association for Computational Linguistics},
  5:135--146.

\bibitem[{Devlin et~al.(2019)Devlin, Chang, Lee, and
  Toutanova}]{devlin2018bert}
Jacob Devlin, Ming-Wei Chang, Kenton Lee, and Kristina Toutanova. 2019.
\newblock \href {https://doi.org/10.18653/v1/N19-1423} {{{BERT}: Pre-training
  of Deep Bidirectional Transformers for Language Understanding}}.
\newblock In \emph{Proceedings of the 2019 Conference of the North {A}merican
  Chapter of the Association for Computational Linguistics: Human Language
  Technologies, Volume 1 (Long and Short Papers)}, pages 4171--4186,
  Minneapolis, Minnesota. Association for Computational Linguistics.

\bibitem[{Dror et~al.(2018)Dror, Baumer, Shlomov, and
  Reichart}]{dror18significance}
Rotem Dror, Gili Baumer, Segev Shlomov, and Roi Reichart. 2018.
\newblock \href {http://aclweb.org/anthology/P18-1128} {{The Hitchhiker's Guide
  to Testing Statistical Significance in Natural Language Processing}}.
\newblock In \emph{Proceedings of the 56th Annual Meeting of the Association
  for Computational Linguistics (Volume 1: Long Papers)}, pages 1383--1392.
  Association for Computational Linguistics.

\bibitem[{Gamb{\"a}ck and Das(2014)}]{gamback2014measuring}
Bj{\"o}rn Gamb{\"a}ck and Amitava Das. 2014.
\newblock \href
  {https://pdfs.semanticscholar.org/c82c/9ea0073129904738fbc051c06188c02f4f6b.pdf?_ga=2.227000350.2022457670.1588360002-52654363.1584622740}
  {{On Measuring the Complexity of Code-Mixing}}.
\newblock In \emph{Proceedings of the 11th International Conference on Natural
  Language Processing, Goa, India}, pages 1--7.

\bibitem[{Gehring et~al.(2017)Gehring, Auli, Grangier, Yarats, and
  Dauphin}]{DBLP:journals/corr/GehringAGYD17}
Jonas Gehring, Michael Auli, David Grangier, Denis Yarats, and Yann~N. Dauphin.
  2017.
\newblock \href {http://arxiv.org/abs/1705.03122} {{Convolutional Sequence to
  Sequence Learning}}.
\newblock \emph{CoRR}, abs/1705.03122.

\bibitem[{Howard and Ruder(2018)}]{howard-ruder-2018-universal}
Jeremy Howard and Sebastian Ruder. 2018.
\newblock \href {https://doi.org/10.18653/v1/P18-1031} {{Universal Language
  Model Fine-tuning for Text Classification}}.
\newblock In \emph{Proceedings of the 56th Annual Meeting of the Association
  for Computational Linguistics (Volume 1: Long Papers)}, pages 328--339,
  Melbourne, Australia. Association for Computational Linguistics.

\bibitem[{Jain and Bhat(2014)}]{jain-bhat-2014-language}
Naman Jain and Riyaz~Ahmad Bhat. 2014.
\newblock \href {https://doi.org/10.3115/v1/W14-3910} {{Language Identification
  in Code-Switching Scenario}}.
\newblock In \emph{Proceedings of the First Workshop on Computational
  Approaches to Code Switching}, pages 87--93, Doha, Qatar. Association for
  Computational Linguistics.

\bibitem[{Mave et~al.(2018)Mave, Maharjan, and
  Solorio}]{mave-etal-2018-language}
Deepthi Mave, Suraj Maharjan, and Thamar Solorio. 2018.
\newblock \href {https://www.aclweb.org/anthology/W18-3206} {{Language
  Identification and Analysis of Code-Switched Social Media Text}}.
\newblock In \emph{Proceedings of the Third Workshop on Computational
  Approaches to Linguistic Code-Switching}, pages 51--61, Melbourne, Australia.
  Association for Computational Linguistics.

\bibitem[{Molina et~al.(2016)Molina, AlGhamdi, Ghoneim, Hawwari,
  Rey-Villamizar, Diab, and Solorio}]{molina-etal-2016-overview}
Giovanni Molina, Fahad AlGhamdi, Mahmoud Ghoneim, Abdelati Hawwari, Nicolas
  Rey-Villamizar, Mona Diab, and Thamar Solorio. 2016.
\newblock \href {https://doi.org/10.18653/v1/W16-5805} {{Overview for the
  Second Shared Task on Language Identification in Code-Switched Data}}.
\newblock In \emph{Proceedings of the Second Workshop on Computational
  Approaches to Code Switching}, pages 40--49, Austin, Texas. Association for
  Computational Linguistics.

\bibitem[{Pennington et~al.(2014)Pennington, Socher, and
  Manning}]{pennington-etal-2014-glove}
Jeffrey Pennington, Richard Socher, and Christopher Manning. 2014.
\newblock \href {https://doi.org/10.3115/v1/D14-1162} {{{G}lo{V}e: Global
  Vectors for Word Representation}}.
\newblock In \emph{Proceedings of the 2014 Conference on Empirical Methods in
  Natural Language Processing ({EMNLP})}, pages 1532--1543, Doha, Qatar.
  Association for Computational Linguistics.

\bibitem[{Peters et~al.(2018)Peters, Neumann, Iyyer, Gardner, Clark, Lee, and
  Zettlemoyer}]{peters-EtAl:2018:N18-1}
Matthew Peters, Mark Neumann, Mohit Iyyer, Matt Gardner, Christopher Clark,
  Kenton Lee, and Luke Zettlemoyer. 2018.
\newblock \href {http://www.aclweb.org/anthology/N18-1202} {{Deep
  Contextualized Word Representations}}.
\newblock In \emph{Proceedings of the 2018 Conference of the North American
  Chapter of the Association for Computational Linguistics: Human Language
  Technologies, Volume 1 (Long Papers)}, pages 2227--2237, New Orleans,
  Louisiana. Association for Computational Linguistics.

\bibitem[{Petrov et~al.(2012)Petrov, Das, and
  McDonald}]{petrov-etal-2012-universal}
Slav Petrov, Dipanjan Das, and Ryan McDonald. 2012.
\newblock \href
  {http://www.lrec-conf.org/proceedings/lrec2012/pdf/274_Paper.pdf} {{A
  Universal Part-of-Speech Tagset}}.
\newblock In \emph{Proceedings of the Eighth International Conference on
  Language Resources and Evaluation ({LREC}'12)}, pages 2089--2096, Istanbul,
  Turkey. European Language Resources Association (ELRA).

\bibitem[{Schwenk(2018)}]{schwenk-2018-filtering}
Holger Schwenk. 2018.
\newblock \href {https://doi.org/10.18653/v1/P18-2037} {{Filtering and Mining
  Parallel Data in a Joint Multilingual Space}}.
\newblock In \emph{Proceedings of the 56th Annual Meeting of the Association
  for Computational Linguistics (Volume 2: Short Papers)}, pages 228--234,
  Melbourne, Australia. Association for Computational Linguistics.

\bibitem[{Schwenk and Douze(2017)}]{schwenk-douze-2017-learning}
Holger Schwenk and Matthijs Douze. 2017.
\newblock \href {https://doi.org/10.18653/v1/W17-2619} {{Learning Joint
  Multilingual Sentence Representations with Neural Machine Translation}}.
\newblock In \emph{Proceedings of the 2nd Workshop on Representation Learning
  for {NLP}}, pages 157--167, Vancouver, Canada. Association for Computational
  Linguistics.

\bibitem[{Schwenk and Li(2018)}]{SCHWENK18.658}
Holger Schwenk and Xian Li. 2018.
\newblock \href {https://www.aclweb.org/anthology/L18-1560} {{A Corpus for
  Multilingual Document Classification in Eight Languages}}.
\newblock In \emph{Proceedings of the Eleventh International Conference on
  Language Resources and Evaluation ({LREC} 2018)}, Miyazaki, Japan. European
  Language Resources Association (ELRA).

\bibitem[{Singh et~al.(2018)Singh, Sen, and
  Kumaraguru}]{singh-etal-2018-twitter}
Kushagra Singh, Indira Sen, and Ponnurangam Kumaraguru. 2018.
\newblock \href {https://doi.org/10.18653/v1/W18-3503} {{A Twitter Corpus for
  {H}indi-{E}nglish Code Mixed {POS} Tagging}}.
\newblock In \emph{Proceedings of the Sixth International Workshop on Natural
  Language Processing for Social Media}, pages 12--17, Melbourne, Australia.
  Association for Computational Linguistics.

\bibitem[{Sitaram et~al.(2019)Sitaram, Chandu, Rallabandi, and
  Black}]{sitaram2019survey}
Sunayana Sitaram, Khyathi~Raghavi Chandu, Sai~Krishna Rallabandi, and Alan~W.
  Black. 2019.
\newblock \href {http://arxiv.org/abs/1904.00784} {{A Survey of Code-switched
  Speech and Language Processing}}.
\newblock \emph{CoRR}, abs/1904.00784.

\bibitem[{Solorio et~al.(2014)Solorio, Blair, Maharjan, Bethard, Diab, Ghoneim,
  Hawwari, AlGhamdi, Hirschberg, Chang, and Fung}]{solorio-etal-2014-overview}
Thamar Solorio, Elizabeth Blair, Suraj Maharjan, Steven Bethard, Mona Diab,
  Mahmoud Ghoneim, Abdelati Hawwari, Fahad AlGhamdi, Julia Hirschberg, Alison
  Chang, and Pascale Fung. 2014.
\newblock \href {https://doi.org/10.3115/v1/W14-3907} {{Overview for the First
  Shared Task on Language Identification in Code-Switched Data}}.
\newblock In \emph{Proceedings of the First Workshop on Computational
  Approaches to Code Switching}, pages 62--72, Doha, Qatar. Association for
  Computational Linguistics.

\bibitem[{Soto and Hirschberg(2018)}]{soto-hirschberg-2018-joint}
Victor Soto and Julia Hirschberg. 2018.
\newblock \href {https://doi.org/10.18653/v1/W18-3201} {{Joint Part-of-Speech
  and Language {ID} Tagging for Code-Switched Data}}.
\newblock In \emph{Proceedings of the Third Workshop on Computational
  Approaches to Linguistic Code-Switching}, pages 1--10, Melbourne, Australia.
  Association for Computational Linguistics.

\bibitem[{Trivedi et~al.(2018)Trivedi, Rangwani, and
  Kumar~Singh}]{trivedi-etal-2018-iit}
Shashwat Trivedi, Harsh Rangwani, and Anil Kumar~Singh. 2018.
\newblock \href {https://www.aclweb.org/anthology/W18-3220} {{{IIT} ({BHU})
  Submission for the {ACL} Shared Task on Named Entity Recognition on
  Code-switched Data}}.
\newblock In \emph{Proceedings of the Third Workshop on Computational
  Approaches to Linguistic Code-Switching}, pages 148--153, Melbourne,
  Australia. Association for Computational Linguistics.

\bibitem[{Vaswani et~al.(2017)Vaswani, Shazeer, Parmar, Uszkoreit, Jones,
  Gomez, Kaiser, and Polosukhin}]{transformers}
Ashish Vaswani, Noam Shazeer, Niki Parmar, Jakob Uszkoreit, Llion Jones,
  Aidan~N Gomez, \L~ukasz Kaiser, and Illia Polosukhin. 2017.
\newblock \href
  {http://papers.nips.cc/paper/7181-attention-is-all-you-need.pdf} {{Attention
  is All you Need}}.
\newblock In I.~Guyon, U.~V. Luxburg, S.~Bengio, H.~Wallach, R.~Fergus,
  S.~Vishwanathan, and R.~Garnett, editors, \emph{Advances in Neural
  Information Processing Systems 30}, pages 5998--6008. Curran Associates, Inc.

\bibitem[{Wang et~al.(2018)Wang, Cho, and Kiela}]{wang-etal-2018-code}
Changhan Wang, Kyunghyun Cho, and Douwe Kiela. 2018.
\newblock \href {https://www.aclweb.org/anthology/W18-3221} {{Code-Switched
  Named Entity Recognition with Embedding Attention}}.
\newblock In \emph{Proceedings of the Third Workshop on Computational
  Approaches to Linguistic Code-Switching}, pages 154--158, Melbourne,
  Australia. Association for Computational Linguistics.

\bibitem[{Winata et~al.(2019)Winata, Lin, and Fung}]{winata-etal-2019-learning}
Genta~Indra Winata, Zhaojiang Lin, and Pascale Fung. 2019.
\newblock \href {https://doi.org/10.18653/v1/W19-4320} {{Learning Multilingual
  Meta-Embeddings for Code-Switching Named Entity Recognition}}.
\newblock In \emph{Proceedings of the 4th Workshop on Representation Learning
  for NLP (RepL4NLP-2019)}, pages 181--186, Florence, Italy. Association for
  Computational Linguistics.

\bibitem[{Winata et~al.(2018)Winata, Wu, Madotto, and
  Fung}]{winata-etal-2018-bilingual}
Genta~Indra Winata, Chien-Sheng Wu, Andrea Madotto, and Pascale Fung. 2018.
\newblock \href {https://www.aclweb.org/anthology/W18-3214} {{Bilingual
  Character Representation for Efficiently Addressing Out-of-Vocabulary Words
  in Code-Switching Named Entity Recognition}}.
\newblock In \emph{Proceedings of the Third Workshop on Computational
  Approaches to Linguistic Code-Switching}, pages 110--114, Melbourne,
  Australia. Association for Computational Linguistics.

\bibitem[{Yirmibe{\c{s}}o{\u{g}}lu and
  Eryi{\u{g}}it(2018)}]{yirmibesoglu-eryigit-2018-detecting}
Zeynep Yirmibe{\c{s}}o{\u{g}}lu and G{\"u}l{\c{s}}en Eryi{\u{g}}it. 2018.
\newblock \href {https://www.aclweb.org/anthology/W18-6115} {{Detecting
  Code-Switching between {T}urkish-{E}nglish Language Pair}}.
\newblock In \emph{Proceedings of the 2018 {EMNLP} Workshop W-{NUT}: The 4th
  Workshop on Noisy User-generated Text}, pages 110--115, Brussels, Belgium.
  Association for Computational Linguistics.

\end{thebibliography}
\bibliographystyle{acl_natbib}

\appendix

\section*{\centering Appendix for ``From English to Code-Switching: Transfer Learning with Strong Morphological Clues''}
\label{sec:supplemental}

\section{Language Identification Distributions}
\label{app:lid-full-distribution}

Table \ref{table:label_stats_lid} shows the distribution of the language identification labels across the CALCS datasets.
\begin{table}[ht!]
\centering
\small
\resizebox{\linewidth}{!}{
\begin{tabular}{lrrr}
\toprule
\textbf{Labels}  & \textbf{Nep-Eng}  & \textbf{Spa-Eng} & \textbf{Hin-Eng} 
\\\midrule
\texttt{lang1}       & 71,148    & 112,579   & 84,752  \\
\texttt{lang2}       & 64,534    & 119,408   & 29,958\\
\texttt{other}       & 45,286    & 55,768    & 21,725\\
\texttt{ne}          & 5,053     & 5,693     & 9,657\\
\texttt{ambiguous}   & 126       & 404       & 13\\
\texttt{mixed}       & 177       & 54        & 58\\
\texttt{fw}          & 0         & 30        & 542\\
\texttt{unk}         & 0         & 325       & 17\\
\bottomrule
\end{tabular}
}
\caption{\label{table:label_stats_lid} Label distribution for LID datasets.} 
\end{table} 

We notice that the CALCS datasets have monolingual tweets, which we detail at the utterance-level in Table \ref{table:utterance_stats_lid}. 
We use the information in this table to measure the rate of code-switching by using the Code-Mixed Index (CMI) \cite{gamback2014measuring}. 
The higher the score of the CMI, the more code-switched the text is. We show the CMI scores in Table \ref{table:cmi}.

\begin{table}[ht!]
\centering
\small
\resizebox{\linewidth}{!}{
    \begin{tabular}{lrrr}
    \toprule
    \textbf{Labels}  & \textbf{Nep-Eng}  & \textbf{Spa-Eng} & \textbf{Hin-Eng} \\
    \midrule
    code-switched     & 9,868    & 8,733 & 3,237  \\
    \texttt{lang1}    & 1,374    & 8,427& 3,842\\
    \texttt{lang2}     & 1,614    & 7,273& 298\\
    \texttt{other}      & 11    & 697& 44\\
    \bottomrule
    \end{tabular}
}
\caption{\label{table:utterance_stats_lid} Utterance level language distribution for language identification datasets.} 
\end{table}

\begin{table}[ht!]
\centering
\small
\resizebox{\linewidth}{!}{
    \begin{tabular}{lll}
    \toprule
    \textbf{Corpus}  & \textbf{CMI-all}  & \textbf{CMI-mixed} 
    \\\midrule
    Nepali-English 2014     & 19.708    & 25.697  \\
    Spanish-English 2016    & 7.685     & 22.114\\
    Hindi-English 2018      & 10.094    & 23.141\\
    \bottomrule
    \end{tabular}
}
\caption{\label{table:cmi}Code-Mixing Index (CMI) for the language identification datasets. CMI-all: average over
all utterances in the corpus. CMI-mixed: average over only code-switched instances.}
\end{table}

\section{Parts-of-Speech Label Distribution}

Table \ref{tab:pos-distribution} shows the distribution of the POS tags for Hindi-English. This dataset correspond to the POS tagging experiments in Section \ref{sec:exp-transfer-learning}.

\label{app:pos-distribution}
\begin{table}[ht!]
\centering
\small
\resizebox{0.75\linewidth}{!}{
    \begin{tabular}{llll}
    \toprule
    \textbf{POS Labels}& \textbf{Train} 	& \textbf{Dev}	& \textbf{Test}\\\midrule
    \texttt{X}					& 5296	& 790	& 1495 \\
    \texttt{VERB}				& 4035	& 669	& 1280 \\
    \texttt{NOUN}				& 3511	& 516	& 1016 \\
    \texttt{ADP}				& 2037	& 346	& 599 \\
    \texttt{PROPN}				& 1996	& 271	& 470 \\
    \texttt{ADJ}				& 1070	& 170	& 308 \\
    \texttt{PART}				& 1045	& 145	& 23 \\
    \texttt{PRON}				& 1013	& 159	& 284 \\
    \texttt{DET}				& 799	& 116	& 226 \\
    \texttt{ADV}				& 717	& 100	& 204 \\
    \texttt{CONJ}				& 571	& 77	& 161 \\
    \texttt{PART\_NEG}			& 333	& 43	& 92 \\
    \texttt{PRON\_WH}			& 294	& 39	& 88 \\
    \texttt{NUM}				& 276	& 35	& 80 \\
    \bottomrule
    \end{tabular} 
}
\caption{The POS tag distribution for Hindi-English.}
\label{tab:pos-distribution}
\end{table}

\section{Named Entity Recognition Label Distribution}
\label{app:ner-distribution}

Table \ref{tab:ner-distribution} shows the distribution of the NER labels for Spanish-English. This dataset corresponds to the NER experiments in Section \ref{sec:exp-transfer-learning}.

\begin{table}[ht!]
\centering
\small
\resizebox{0.9\linewidth}{!}{
\begin{tabular}{llll}
\toprule
\textbf{NER Classes}& \textbf{Train} 	& \textbf{Dev}	& \textbf{Test}\\\midrule
\texttt{person} 			& 6,226				& 95			& 1,888 	\\
\texttt{location} 			& 4,323				& 16			& 803   	\\
\texttt{organization}		& 1,381				& 10			& 307   	\\
\texttt{group}	     		& 1,024				& 5				& 153   	\\
\texttt{title}				& 1,980				& 50			& 542   	\\
\texttt{product}			& 1,885				& 21			& 481   	\\
\texttt{event}				& 557				& 6				& 99    	\\
\texttt{time}				& 786				& 9				& 197   	\\
\texttt{other}				& 382				& 7				& 62    	\\\midrule
NE Tokens			        & 18,544			& 219 			& 4,532		\\
O Tokens 			        & 614,013			& 9,364			& 178,479 	\\\midrule
Tweets 				        & 50,757			& 832			& 15,634	\\\bottomrule
\end{tabular}
}
\caption{The distribution of labels for the Spanish-English NER dataset from CALCS 2018.}
\label{tab:ner-distribution}
\end{table}

\section{Hyperparameters and Fine-tuning}
\label{app:sec:fine-tuning}

\begin{figure*}[t!]
\centering
\includegraphics[width=\linewidth]{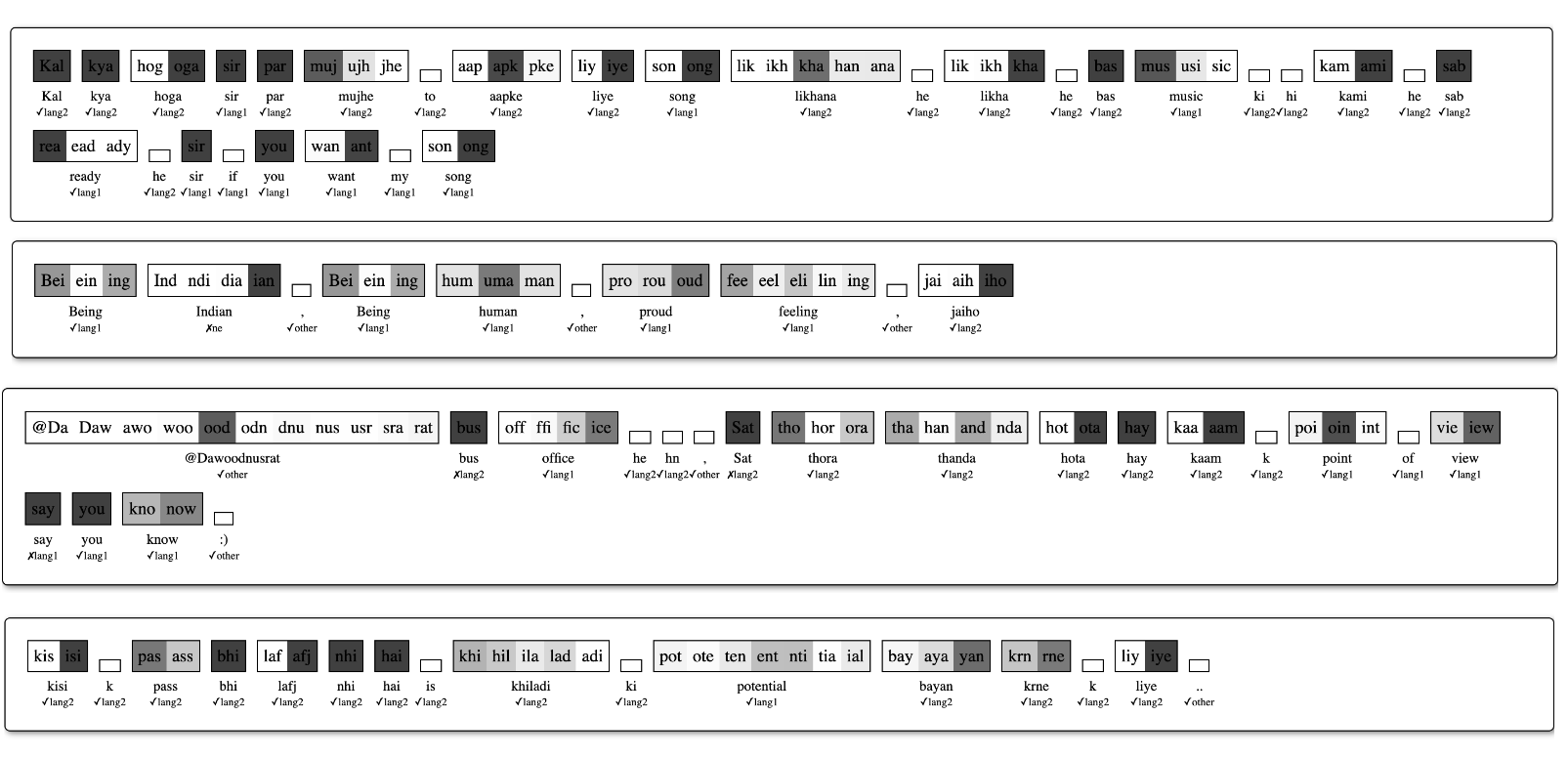}
\caption{ Visualization of the attention weights at the tri-gram level for the Hindi-English 2018 dataset on the LID task. The boxes contain the tri-grams of the word below them. We also provide the predicted label by the model, and whether it was correct or wrong.}
\label{app:fig:attention-visualization}
\end{figure*}

We experiment with our LID models using Adam optimizer with a learning rate of $0.001$ and a plateau learning rate scheduler with patience of 5 epochs based on the validation loss. We train our LID models using this setting for 50 epochs. For the last block of experiments in Table \ref{tab:lid-experiments}, we use a progressive fine-tuning process described below.

\paragraph{Fine-tuning.} 
We fine-tune the model by progressively updating the parameters from the top to the bottom layers of the model. 
This avoids losing the pre-trained knowledge from ELMo and smoothly adapts the network to the new languages from the code-switched data. 
We use the slanted triangular learning rate scheduler with both gradual unfreezing and discriminative fine-tuning over the layers (i.e., different learning rates across layers) proposed by \citet{howard-ruder-2018-universal}. 
We group the non-ELMo parameters of our model apart from the ELMo parameters. 
We set the non-ELMo parameters to be the first group of parameters to be tuned (i.e., parameters from enhanced character n-grams, CRF, and BLSTM). Then, we further group the ELMo parameters as follows (top to bottom):
\begin{enumerate}
    \item the second bidirectional LSTM layer, 
    \item the first bidirectional LSTM layer, 
    \item the highway network, 
    \item the linear projection from flattened convolutions to the token embedding space, 
    \item all the convolutional layers, and 6) the character embedding weights. 
\end{enumerate}
Once all the layers have been unfrozen, we update all the parameters together. This technique allows us get the most of our model moving from English to a code-switching setting. We train our fine-tuned models for 200 epochs and a initial learning rate of 0.01 that gets modified during training.

Additionally, we use this fine-tuning process for the downstream NLP task presented in the paper (i.e., NER and POS tagging).

\section{Visualization of Attention Weights for Hindi-English}
\label{sec:appendix_hi_en_viz}

Figure \ref{app:fig:attention-visualization} shows the attention behavior for tri-grams on the Hindi-English dataset. 
Similar to the cases discussed for Spanish-English in the main content, we observe that the model learns tri-grams like \textit{-ing, -ian} for English and \textit{iye, isi} for Hindi.
\end{document}